\newif\ifllncs
\llncsfalse
\ifllncs
\documentclass[10pt]{llncs}
\usepackage{makeidx}
\else
\documentclass[10pt,twocolumn]{article}
\fi
\usepackage{natbib}
\usepackage[english]{babel}
\usepackage[utf8]{inputenc}
\usepackage{amsmath}
\usepackage{algorithm}
\usepackage{algorithmic}

\begin{document}

\def\mytitle{Practical Recommendations for Gradient-Based Training of Deep Architectures} 

\ifllncs
\frontmatter
\pagestyle{headings}  
\addtocmark{\mytitle} 
\fi

\author{Yoshua Bengio}
\ifllncs
\institute{Université de Montréal}
\fi
\title{\mytitle}
\date{Version 2, Sept. 16th, 2012}

\maketitle

\begin{abstract}
  Learning algorithms related to artificial neural networks and in particular for Deep Learning may seem to involve many bells and whistles, called hyper-parameters. This chapter is meant as a practical guide with recommendations for some of the most commonly used hyper-parameters, in particular in the context of learning algorithms based on back-propagated gradient and gradient-based optimization. It also discusses how to deal with the fact that more interesting results can be obtained when allowing one to adjust many hyper-parameters. Overall, it describes elements of the practice used to successfully and efficiently train and debug large-scale and often deep multi-layer neural networks. It closes with open questions about the training difficulties observed with deeper architectures.
\end{abstract}


\section{Introduction}



Following a decade of lower activity, research in artificial neural networks was 
revived after a 2006 breakthrough~\citep{Hinton06-small,Bengio-nips-2006-small,ranzato-07-small} 
in the area of {\em Deep Learning}, based on greedy layer-wise unsupervised
pre-training of each layer of features. See~\citep{Bengio-2009-book} for a review.
Many of the practical recommendations that
justified the previous edition of this book are still valid,
and new elements were added, while some survived longer
by virtue of the practical advantages they provided. The
panorama presented in this chapter regards some of these surviving
or novel elements of practice, focusing on learning algorithms aiming at training
deep neural networks, but leaving most of the material specific to the Boltzmann machine
family to another chapter~\citep{Hinton-tricks-chapter}.

Although such recommendations come out of a living
practice that emerged from years of experimentation and to 
some extent mathematical justification, they should be
challenged. They constitute a good starting point for the
experimenter and user of learning algorithms but very often have not been formally
validated, leaving open many questions that can be
answered either by theoretical analysis or by solid comparative
experimental work (ideally by both). A good indication
of the need for such validation is that different researchers
and research groups do not always agree on the practice
of training neural networks.

Several of the recommendations presented here can be
found implemented in the {\em Deep Learning 
Tutorials}\footnote{~\tt http://deeplearning.net/tutorial/} and
in the related {\em Pylearn2} library\footnote{~\tt http://deeplearning.net/software/pylearn2}, all based on
the {\tt Theano} library (discussed below) written in the {\em Python}
programming language.

The 2006 Deep Learning 
breakthrough~\citep{Hinton06-small,Bengio-nips-2006-small,ranzato-07-small} 
centered on the use of {\em unsupervised representation
  learning} to help learning {\em internal 
representations}\footnote{~A neural network computes a sequence of
data transformations, each step encoding the raw input into an intermediate
or internal representation, in principle to make the prediction or modeling 
task of interest easier.} by providing a
{\em local training signal} at each level of a hierarchy of 
features\footnote{~In standard multi-layer neural networks trained using
back-propagated gradients, the only signal that drives parameter updates
is provided at the output of the network (and then propagated backwards).
Some unsupervised learning algorithms provide a local source of guidance
for the parameter update in each layer, based only on the inputs and outputs
of that layer.}. Unsupervised
representation learning algorithms can be applied several times to learn
different layers of a deep model. Several
unsupervised representation learning algorithms have been proposed since then.
Those covered in this chapter (such as auto-encoder variants) retain many
of the properties of artificial multi-layer neural networks, relying on the
back-propagation algorithm to estimate stochastic gradients. 
Deep Learning algorithms such
as those based on the Boltzmann machine and those based on
auto-encoder or sparse coding variants
often include a supervised fine-tuning stage. This supervised fine-tuning as
well as the gradient descent performed with auto-encoder variants also involves
the back-propagation algorithm, just as like when training deterministic
feedforward or recurrent artificial neural networks.
Hence this chapter also includes recommendations for training ordinary
supervised deterministic neural networks or more generally,
most machine learning algorithms relying on iterative gradient-based optimization
of a parametrized learner with respect to an explicit training criterion.

This chapter assumes that the reader already understands the standard
algorithms for training supervised multi-layer neural networks, with
the loss gradient computed thanks to the back-propagation 
algorithm~\citep{Rumelhart86b}. It starts by explaining 
basic concepts behind Deep Learning and the greedy layer-wise
pretraining strategy (Section~\ref{sec:deep}), and recent
unsupervised pre-training algorithms (denoising and contractive
auto-encoders) that are closely related in
the way they are trained to standard multi-layer neural networks
(Section~\ref{sec:ae}). It then reviews in Section~\ref{sec:gradients} 
basic concepts in
iterative gradient-based optimization and in particular the
stochastic gradient method, gradient computation with a flow graph,
automatic differentation.
The main section of this chapter is Section~\ref{sec:hyper},
which explains hyper-parameters in general, their optimization,
and specifically covers the main hyper-parameters of neural networks.
Section~\ref{sec:debug} briefly describes simple ideas and methods
to debug and visualize neural networks, while Section~\ref{sec:other}
covers parallelism, sparse high-dimensional inputs, symbolic
inputs and embeddings, and multi-relational learning. The chapter
closes (Section~\ref{sec:open}) with open questions on the difficulty
of training deep architectures and improving the optimization methods
for neural networks.

\subsection{Deep Learning and Greedy Layer-Wise Pretraining}
\label{sec:deep}

The notion of {\em reuse}, which explains the power of distributed
representations~\citep{Bengio-2009-book}, is also at the heart of the
theoretical advantages behind {\em Deep Learning}. Complexity theory of
circuits, e.g.~\citep{Hastad86-small,Hastad91}, (which include neural
networks as special cases) has much preceded the recent research on deep
learning. The depth of a circuit is the length of the longest path from an
input node of the circuit to an output node of the circuit.  Formally, one
can change the depth of a given circuit by changing the definition of what
each node can compute, but only by a constant
factor~\citep{Bengio-2009-book}. The typical computations we allow in each
node include: weighted sum, product, artificial neuron model (such as a
monotone non-linearity on top of an affine transformation), computation of
a kernel, or logic gates.  Theoretical
results~\citep{Hastad86-small,Hastad91,Bengio-localfailure-NIPS-2006-small,Bengio+Lecun-chapter2007-small,Bengio+Delalleau-ALT-2011-small}
clearly identify families of functions where a deep representation can be
exponentially more efficient than one that is insufficiently deep. If the
same set of functions can be represented from within a family of
architectures associated with a smaller
VC-dimension (e.g. less hidden units\footnote{~Note that in our
experiments, deep architectures tend to generalize very well even when they
have quite large numbers of parameters.}), learning theory would suggest
that it can be learned with fewer examples, yielding improvements in both 
{\em computational} efficiency and {\em statistical} efficiency.

Another important motivation for feature learning and Deep Learning is that
they can be done with {\em unlabeled examples}, so long as the factors (unobserved random variables
explaining the data)
relevant to the questions we will ask later (e.g. classes to be predicted)
are somehow salient in the input
distribution 
itself. This is true under the {\em manifold hypothesis}, which states that
natural classes and other high-level concepts in which humans are
interested 
are associated with {\em low-dimensional} regions in input space (manifolds) near
which the distribution concentrates, and that different class manifolds are
well-separated by regions of very low density. It means that a small semantic
change around a particular example can be captured by changing only a few numbers in a high-level
abstract representation space.
As a consequence, feature
learning and Deep Learning are intimately related to principles of {\em
  unsupervised learning}, and they can work in the {\em
  semi-supervised setting} (where only a few examples are labeled), as well
as in the {\em transfer learning} and {\em multi-task} settings (where we aim
to generalize to new classes or tasks). 
The underlying hypothesis is that many of the underlying factors are {\em
  shared} across classes or tasks. Since representation learning aims to
extract and isolate these factors, representations can be {\em shared}
across classes and tasks.   

One of the most commonly used approaches for
training deep neural networks is based on {\em greedy layer-wise
  pre-training}~\citep{Bengio-nips-2006-small}. The idea, first introduced
in~\citet{Hinton06-small}, is to train one layer of a deep architecture at a time 
using unsupervised representation learning. Each level takes as input the
representation learned at the previous level and learns a new
representation. The learned representation(s) can then be used as input to
predict variables of interest, for example to classify objects. After unsupervised
pre-training, one can
also perform supervised fine-tuning of the whole system\footnote{~The whole
system composes the computation of the
representation with computation of the predictor's output.}, i.e., optimize
not just the classifier but also the lower levels of the feature hierarchy
with respect to some objective of interest.  Combining unsupervised pre-training
and supervised fine-tuning usually gives better generalization than pure
supervised learning from a purely random initialization.
The unsupervised
representation learning algorithms for pre-training
proposed in 2006 were the Restricted Boltzmann
Machine or RBM~\citep{Hinton06-small}, the auto-encoder~\citep{Bengio-nips-2006-small}
and a sparsifying form of auto-encoder similar to sparse
coding~\citep{ranzato-07-small}.

\subsection{Denoising and Contractive Auto-Encoders}
\label{sec:ae}

An auto-encoder has two parts: an encoder function $f$ that maps
the input $x$ to a representation $h=f(x)$, and a decoder function
$g$ that maps $h$ back in the space of $x$ in order to reconstruct
$x$. In the regular auto-encoder the reconstruction function $r(\cdot)=g(f(\cdot))$
is trained to minimize the average value of a reconstruction loss {\em
on the training examples}. Note that reconstruction loss should be high for 
most other input configurations\footnote{~Different regularization mechanisms have been
proposed to push reconstruction error up in low density areas: denoising
criterion, contractive criterion, and code sparsity. It has been argued
that such constraints play a role similar to the partition function
for Boltzmann machines~\citep{ranzato-08}.}. The regularization mechanism
makes sure that reconstruction cannot be perfect everywhere, while minimizing
the reconstruction loss at training examples digs a hole in reconstruction
error where the density of training examples is large. Examples of 
reconstruction loss functions include
$||x - r(x)||^2$ (for real-valued inputs) and
$-\sum_i x_i \log r_i(x) + (1-x_i) \log (1-r_i(x))$ (when interpreting
$x_i$ as a bit or a probability of a binary event). Auto-encoders
capture the input distribution by learning to {\em better reconstruct
more likely input configurations}. The difference between the reconstruction
vector and the input vector can be shown to be related to the
log-density gradient as estimated by the 
learner~\citep{Vincent-NC-2011-small,Bengio-arxiv-moments-2012}
and the Jacobian matrix of the reconstruction with respect to the input
gives information about the second derivative of the density, i.e.,
in which direction the density remains high when you are on a high-density
manifold~\citep{Rifai+al-2011-small,Bengio-arxiv-moments-2012}.
In the
Denoising Auto-Encoder (DAE) and the Contractive Auto-Encoder (CAE),
the training procedure also introduces {\em robustness} (insensitivity
to small variations), respectively in the reconstruction $r(x)$ or in
the representation $f(x)$. In the 
DAE~\citep{VincentPLarochelleH2008-small,Vincent-JMLR-2010-small}, this is achieved
by training with stochastically corrupted inputs, but trying to reconstruct
the uncorrupted inputs. In the CAE~\citep{Rifai+al-2011-small}, this is
achieved by adding an explicit regularizing term in the training
criterion, proportional to the norm of the Jacobian of the
encoder, $||\frac{\partial f(x)}{\partial x}||^2$. But the CAE and the
DAE are very related~\citep{Bengio-arxiv-moments-2012}:
when the noise is Gaussian and small, the denoising error minimized
by the DAE is equivalent
to minimizing the norm of the Jacobian of the reconstruction
function $r(\cdot)=g(f(\cdot))$, whereas the CAE minimizes the norm of the Jacobian of
the encoder $f(\cdot)$. Besides Gaussian noise, another interesting form of corruption has
been very successful with DAEs: it is called the {\em masking corruption} and consists
in randomly zeroing out a large fraction (like 20\% or even 50\%) of the inputs, where
the zeroed out subset is randomly selected for each example. In addition to the contractive
effect, it forces the learned encoder to be able to rely only on an arbitrary subset of
the input features.

Another way to prevent the auto-encoder
from perfectly reconstructing everywhere is to introduce a sparsity
penalty on $h$, discussed below (Section~\ref{sec:nnets}).

\subsection{Online Learning and Optimization of Generalization Error}

The objective of learning is not to minimize training error or even
the training criterion. The latter is a surrogate for generalization
error, i.e., performance on new (out-of-sample) examples, and there are
no hard guarantees that minimizing the training criterion will yield
good generalization error: it depends on the appropriateness of the
parametrization and training criterion (with the corresponding prior
they imply) for the task at hand.

Many learning tasks of interest will require huge quantities of data
(most of which will be unlabeled) and as the number of examples increases, so long as capacity
is limited (the number of parameters is small compared to the number of
examples), training error and generalization approach each other.
In the regime of such large datasets, we can 
consider that the learner sees an unending stream of examples (e.g.,
think about a process that harvests text and images from the web and
feeds it to a machine learning algorithm). In that context, it is most
efficient to simply update the parameters of the model after each example
or few examples, as they arrive. This is the ideal {\em online learning}
scenario, and in a simplified setting, we can even consider each new
example $z$ as being sampled i.i.d. from an unknown generating distribution with probability density $p(z)$. 
More realistically, examples in online
learning do not arrive i.i.d. but instead from an unknown stochastic
process which exhibits serial correlation and other temporal dependencies. 
Many learning algorithms rely on gradient-based numerical optimization of
a training criterion. Let $L(z,\theta)$ be the loss incurred on
example $z$ when the parameter vector takes value $\theta$.
The gradient vector for the loss associated with a single
example is $\frac{\partial L(z,\theta)}{\partial \theta}$.

If we consider the simplified case of i.i.d. data, there
is an interesting observation to be made: {\em the online learner
is performing stochastic gradient descent on its generalization error.}
Indeed, the generalization error $C$ of a learner with parameters $\theta$ 
and loss function $L$ is 
\[
   C=E[ L (z, \theta) ] = \int p(z) L(z,\theta) dz
\]
while the stochastic gradient from sample $z$ is
\[
   \hat{g} = \frac{\partial L(z,\theta)}{\partial \theta}
\]
with $z$ a random variable sampled from $p$.
The gradient of generalization error is
\[
  \frac{\partial C}{\partial \theta} = \frac{\partial}{\partial \theta} \int p(z) L(z,\theta) dz
                                     = \int p(z) \frac{\partial L(z,\theta)}{\partial \theta} dz
                                     = E[\hat{g}]
\]
showing that the online gradient $\hat{g}$ is an unbiased estimator of the
generalization error gradient $\frac{\partial C}{\partial \theta}$.  {\em
  It means that online learners, when given a stream of non-repetitive
  training data, really optimize (maybe not in the optimal way, i.e., using
  a first-order gradient technique) what we really care about:
  generalization error.}

\section{Gradients}
\label{sec:gradients}

\subsection{Gradient Descent and Learning Rate}

The gradient or an estimator of the gradient 
is used as the core part the computation of parameter updates for
gradient-based numerical optimization algorithms. For example, 
simple online (or stochastic) 
gradient descent~\citep{robbins_monro:1951,bottou-lecun-04b-small} updates the parameters after each
example is seen, according to
\[
  \theta^{(t)} \leftarrow \theta^{(t-1)} - \epsilon_t \frac{\partial L(z_t,\theta)}{\partial \theta}
\]
where $z_t$ is an example sampled at iteration $t$ and where $\epsilon_t$
is a {\em hyper-parameter} that is called the {\em learning rate} and whose
choice is crucial. If the learning rate is too large\footnote{~above a
value which is approximately 2 times the largest eigenvalue of the average
loss Hessian matrix}, the average loss will increase. The optimal learning
rate is usually close to (by a factor of 2) the {\em largest learning rate that does not cause
divergence of the training criterion}, an observation that can guide
heuristics for setting the learning
rate~\citep{UTLC+DL+tutorial-2011-small}, e.g., start with a large learning
rate and if the training criterion diverges, try again with 3 times smaller
learning rate, etc., until no divergence is observed.

See~\cite{Bottou-tricks-chapter-2013} for a deeper treatment of stochastic gradient descent,
including suggestions to set learning rate schedule and improve the asymptotic convergence
through averaging.

In practice, we use {\em mini-batch} updates based on an {\em average} of the 
gradients\footnote{~Compared 
to a sum, an average makes a small change in $B$ have only a small effect on the optimal learning rate, with
an increase in $B$ generally allowing a small increase in the learning rate because of the reduced
variance of the gradient.}
inside each block of $B$ examples:
\begin{equation}
  \theta^{(t)} \leftarrow \theta^{(t-1)} - 
  \epsilon_t \frac{1}{B}\sum_{t'=Bt+1}^{B(t+1)} \frac{\partial L(z_{t'},\theta)}{\partial \theta}.
\label{eq:minibatch-update}
\end{equation}
With $B=1$ we are back to ordinary online gradient descent, while with $B$ equal to the
training set size, this is standard (also called ``batch'') gradient descent. With intermediate
values of $B$ there is generally a sweet spot. When $B$ increases
we can get more multiply-add operations per second by taking advantage of parallelism or efficient
matrix-matrix multiplications (instead of separate matrix-vector multiplications), often gaining
a factor of 2 in practice in overall training time. On the other hand, as $B$ increases,
the number of updates per computation done decreases, which slows down convergence 
(in terms of error vs number of multiply-add operations performed) because
less updates can be done in the same computing time. Combining these two opposing
effects yields a typical U-curve with a sweet spot at an intermediate value of $B$.

Keep in mind that even the true gradient direction (averaging over the whole training set)
is only the steepest descent direction locally but may not point in the right direction
when considering larger steps. In particular, because the training criterion is not quadratic
in the parameters, as one moves in parameter space the optimal descent direction keeps changing.
Because the gradient direction is not quite the right direction of descent, there is no
point in spending a lot of computation to estimate it precisely for gradient descent.
Instead, doing more updates more frequently helps to explore more and faster, especially
with large learning rates. In addition, smaller values of $B$ may benefit from more
exploration in parameter space and a form of regularization both due to the ``noise''
injected in the gradient estimator, which may explain the better test results sometimes
observed with smaller $B$.

When the training set is finite, training proceeds by sweeps through
the training set called an {\em epoch}, and full training usually
requires many epochs (iterations through the training set).
Note that stochastic gradient (either one example at a time or with mini-batches)
is different from ordinary {\em gradient descent}, sometimes called ``batch
gradient descent'', which corresponds to the
case where $B$ equals the training set size, i.e., there is one parameter
update per epoch).
The great advantage of stochastic gradient descent and other
online or minibatch update methods is that their convergence
does not depend on the size of the training set, only on the
number of updates and the richness of the training distribution.
In the limit of a large or infinite training
set, a batch method (which updates only after seeing all the examples)
is hopeless. In fact, even for ordinary datasets of tens or hundreds of thousands
of examples (or more!), stochastic gradient descent converges
much faster than ordinary (batch) gradient descent, and beyond some dataset sizes
the speed-up is almost linear (i.e., doubling the size almost doubles the gain)\footnote{~On
the other hand, batch methods can be parallelized easily, which becomes an important advantage
with currently available forms of computing power.}.
It is really important to use the stochastic version in order to get reasonable
clock-time convergence speeds.

As for any stochastic gradient descent
method (including the mini-batch case), it is important for
efficiency of the estimator that each example or mini-batch
be sampled approximately independently. Because random access
to memory (or even worse, to disk) is expensive, a good approximation,
called incremental gradient~\citep{Bertsekas-2010},
is to visit the examples (or mini-batches) in a fixed order corresponding
to their order in memory or disk
(repeating the examples in the same order on a second epoch, if 
we are not in the pure online case
where each example is visited only once). In this context,
it is safer if the examples or mini-batches are first put in a random order
(to make sure this is the case, it could be useful to first
shuffle the examples). Faster convergence has been observed
if the order in which the mini-batches are visited is changed for
each epoch, which can be reasonably 
efficient if the training set holds in computer memory.
 
\subsection{Gradient Computation and Automatic Differentiation}

The gradient can be either computed manually or through automatic
differentiation. Either way, it helps to structure this computation
as  a {\em flow graph}, in order to prevent mathematical
mistakes and make sure an implementation is computationally efficient.
The computation of the loss $L(z,\theta)$ as a function of $\theta$
is laid out in a graph whose nodes correspond to elementary operations
such as addition, multiplication, and non-linear
operations such as the neural networks activation function (e.g.,
sigmoid or hyperbolic tangent), possibly at the level of
vectors, matrices or tensors. 
The flow graph is directed and acyclic and has three types of nodes:
input nodes, internal nodes, and output nodes.
Each of its nodes is
associated with a numerical output which is the result of the
application of that computation (none in the case of input nodes), 
taking as input the output
of previous nodes in a directed acyclic graph. Example $z$ and 
parameter vector $\theta$
(or their elements) are the input nodes of the graph (i.e., they do not
have inputs themselves) and $L(z,\theta)$ is a scalar
output of the graph. Note that here, in the supervised case, 
$z$ can include an input
part $x$ (e.g. an image) and a target part $y$ (e.g. a target class
associated with an object in the image). In the unsupervised
case $z=x$. In a semi-supervised case, there is a mix of labeled
and unlabeled examples, and $z$ includes $y$
on the labeled examples but not on the unlabeled ones.

In addition to associating a numerical output $o_a$ to each node $a$ of the
flow graph, we can associate a gradient $g_a=\frac{\partial L(z,\theta)}{\partial o_a}$.
The gradient will be defined and computed recursively in the graph, in the
opposite direction of the computation of the nodes' outputs, i.e., whereas
$o_a$ is computed using outputs $o_p$ of {\em predecessor} nodes $p$ of $a$, $g_a$
will be computed using the gradients $g_s$ of {\em successor} nodes $s$ of $a$.
More precisely, the chain rule dictates
\[
  g_a = \sum_s g_s \frac{\partial o_s}{\partial o_a}
\]
where the sum is over immediate successors of $a$. Only output nodes have no
successor, and in particular for the output node that computes $L$, the gradient is set to 1
since $\frac{\partial L}{\partial L}=1$, thus initializing the recursion.
Manual or automatic differentiation then only requires to define
the partial derivative associated with each type of operation 
performed by any node of the graph. 
When implementing gradient descent algorithms with
manual differentiation the result tends to be verbose, brittle code
that lacks modularity -- all bad things in terms of software
engineering. A better approach is to express the flow graph in terms
of objects that modularize how to compute outputs from inputs as well
as how to compute the partial derivatives necessary for gradient
descent.
One can pre-define the
operations of these objects (in a ``forward propagation'' or {\tt fprop} method) and 
their partial derivatives (in a ``backward propagation'' or {\tt bprop} method) and encapsulate these
computations in an object that knows how to compute its output
given its inputs, and how to compute the gradient with respect
to its inputs given the gradient with respect to its output.
This is the strategy adopted in the {\tt Theano} 
library\footnote{~\tt http://deeplearning.net/software/theano/} with its
{\tt Op} objects~\citep{bergstra+al:2010-scipy-small}, as well
as in libraries such as {\tt Torch}\footnote{~\tt http://www.torch.ch}~\citep{Torch-2011}
and {\tt Lush}\footnote{~\tt http://lush.sourceforge.net}.

Compared to {\tt Torch} and {\tt Lush}, {\tt Theano} adds an interesting
ingredient which makes it a full-fledged automatic differentiation tool:
symbolic computation. The flow graph itself (without the numerical values
attached) can be viewed as a symbolic representation (in a data structure) 
of a numerical computation. In {\tt Theano}, the gradient computation
is first performed symbolically, i.e., each {\tt Op} object knows how to create
other {\tt Ops} corresponding to the computation of the partial derivatives
associated with that {\tt Op}. Hence the {\em symbolic differentiation} of the output of
a flow graph with respect to any or all of its input nodes can be performed easily in most cases,
yielding another flow graph which specifies how to compute these gradients,
given the input of the original graph. Since the gradient graph typically
contains the original graph (mapping parameters to loss) as a sub-graph, in order to make computations efficient
it is important to automate (as done in {\tt Theano}) a number of
{\em simplifications} which are graph transformations preserving the
semantics of the output (given the input) but yielding smaller (or more numerically stable
or more efficiently computed) graphs
(e.g., removing redundant computations). To take advantage of the fact that
computing the loss gradient includes as a first step computing the loss itself,
it is advantageous to structure the code so that both the loss and its gradient
are computed at once, with a single graph having multiple outputs.
The advantages of performing
gradient computations symbolically are numerous. First of all, one
can readily compute gradients over gradients, i.e., second derivatives,
which are useful for some learning algorithms.
Second, one can define algorithms or training criteria involving
gradients themselves, as required for example in the Contractive
Auto-Encoder (which uses the norm of a Jacobian matrix in its training
criterion, i.e., really requires second derivatives, which here are
cheap to compute). Third, it makes it easy to implement other useful
graph transformations such as graph simplifications or numerical
optimizations and transformations that help making the numerical
results more robust and more efficient (such as working in the domain of logarithms
of probabilities rather than in the domain of probabilities directly).
Other potential beneficial applications of such symbolic manipulations include
parallelization and additional differential operators (such as the R-operator,
recently implemented in {\tt Theano}, which is very
useful to compute the product of a Jacobian matrix 
$\frac{\partial f(x)}{\partial x}$ or Hessian matrix $\frac{\partial^2 L(x,\theta)}{\partial \theta^2}$
with a vector without ever having to actually compute and store the matrix
itself~\citep{Pearlmutter-1994}).


\section{Hyper-Parameters}
\label{sec:hyper}

A pure learning algorithm can be seen as a function
taking training data as input and producing as output
a function (e.g. a predictor) or model (i.e. a bunch
of functions). However, in practice, 
many learning algorithms involve hyper-parameters, i.e., annoying
knobs to be adjusted. In many algorithms such as Deep Learning algorithms
the number of hyper-parameters (ten or more!) can make the idea of having to adjust all of them
unappealing. In addition, it has been shown that the use of computer clusters for hyper-parameter selection
can have an important effect on results~\citep{Pinto-2009}. Choosing hyper-parameter values is formally equivalent
to the question of {\em model selection}, i.e., given a family or set of learning
algorithms, how to pick the most appropriate one inside the set?
{\em We define a hyper-parameter for a learning algorithm A as a variable to be set
prior to the actual application of A to the data, one that is not directly selected
by the learning algorithm itself}. It is basically an outside control knob.
It can be discrete (as in model selection) or continuous (such as the
learning rate discussed above).
Of course, one can hide these hyper-parameters by wrapping another learning
algorithm, say B, around A, to selects A's hyper-parameters
(e.g. to minimize validation set error). We can then call B a hyper-learner,
and if B has no hyper-parameters itself then the composition of B over A could be
a ``pure'' learning algorithm, with no hyper-parameter. In the end, to apply
a learner to training data, one has to have a pure learning algorithm. The hyper-parameters
can be fixed by hand or tuned by an algorithm, but their value has to be selected.
The value of some hyper-parameters can be selected based on the performance of A on
its training data, but most cannot. For any hyper-parameter that has an impact
on the effective capacity of a learner, it makes more sense to select its
value based on out-of-sample data (outside the training set), e.g., a validation set performance,
online error, or cross-validation error. Note that some learning algorithms
(in particular unsupervised learning algorithms such as 
algorithms for training RBMs by approximate maximum likelihood)
are problematic in this respect because we cannot directly measure the
quantity that is to be optimized (e.g. the likelihood) because it is
intractable. On the other hand, the expected denoising reconstruction
error is easy to estimate (by just averaging the denoising error over
a validation set). 

Once some out-of-sample
data has been used for selecting hyper-parameter values, it cannot be used
anymore to obtain an unbiased estimator of generalization performance,
so one typically uses a test set (or double 
cross-validation\footnote{~Double cross-validation applies recursively the idea of cross-validation,
using an outer loop cross-validation to evaluate generalization error
and then applying an inner loop cross-validation inside each outer loop split's 
training subset (i.e., splitting it again into training
and validation folds) in order to select hyper-parameters for
that split.}, in the
case of small datasets) to estimate generalization error of the
pure learning algorithm (with hyper-parameter selection hidden inside).

\subsection{Neural Network Hyper-Parameters}
\label{sec:nnets}

Different learning algorithms involve different sets of hyper-parameters,
and it is useful to get a sense of the kinds of choices that practitioners
have to make in choosing their values. We focus here mostly on those
relevant to neural networks and Deep Learning algorithms.

\subsubsection{Hyper-Parameters of the Approximate Optimization}

First of all, several learning algorithms can be viewed as the combination
of two elements: a training criterion and a model (e.g., a family of functions,
a parametrization) on the one hand, and on the other hand, a particular
procedure for approximately optimizing this criterion.
Correspondingly, one should distinguish hyper-parameters associated
with the optimizer from hyper-parameters associated with the
model itself, i.e., typically the function class, regularizer
and loss function. We have already mentioned above some of the 
hyper-parameters typically associated with gradient-based optimization.
Here is a more extensive descriptive list,
focusing on those used in stochastic (mini-batch) gradient descent
(although number of training iterations is used for all iterative
optimization algorithms).
\begin{itemize}
\item The {\bf initial learning rate} ($\epsilon_0$ below, Eq.\eqref{eq:lrate}). 
This is often the single most important hyper-parameter and one should
always make sure that it has been tuned (up to approximately a 
factor of 2). Typical values for a neural network with standardized
inputs (or inputs mapped to the (0,1) interval) are less than 1
and greater than $10^{-6}$ but these should not be taken as strict ranges
and greatly depend on the parametrization of the model. A default value 
of 0.01 typically works for standard multi-layer neural networks but
it would be foolish to rely exclusively on this default value. If there is only
time to optimize one hyper-parameter and one uses stochastic gradient 
descent, then this is the hyper-parameter that is worth tuning.
\item The choice of strategy for decreasing or adapting
the {\bf learning rate schedule} (with hyper-parameters such as 
the time constant $\tau$ in Eq.~\eqref{eq:lrate} below).
The default value of \mbox{$\tau\rightarrow\infty$} means that the learning rate is constant over training iterations.
In many cases the benefit of choosing other than this default value is small.
An example of $O(1/t)$ learning rate schedule, used in~\citet{Bergstra+Bengio-2012-small} is 
\begin{equation}
  \epsilon_t = \frac{\epsilon_0 \tau}{\max(t,\tau)}
\label{eq:lrate}
\end{equation}
which keeps the learning rate constant for the first $\tau$ steps and then
decreases it in $O(1/t^\alpha)$, with traditional recommendations (based
on asymptotic analysis of the convex case) suggesting $\alpha=1$. 
See~\citet{Bach+Moulines-2011} for a recent analysis of the rate of convergence
for the general case of $\alpha\leq 1$, suggesting that smaller values
of $\alpha$ should be used in the non-convex case, especially when
using a gradient averaging or momentum technique (see below).
An adaptive and heuristic way of automatically setting $\tau$ above
is to keep $\epsilon_t$ constant until the training criterion stops
decreasing significantly (by more than some relative improvement threshold) 
from epoch to epoch. That threshold is a less sensitive hyper-parameter than $\tau$
itself. An alternative to a fixed schedule with a couple of (global) free hyper-parameters
like in the above formula is the use of an {\em adaptive} learning rate heuristic,
e.g., the simple procedure proposed in ~\cite{Bottou-tricks-chapter-2013}: at regular intervals
during training, using a fixed small subset of the training set (what matters is only
the number of examples used, not what fraction of the whole training set it represents),
continue training with $N$ different choices of learning rate (all in parallel), and
keep the value that gave the best results until the next re-estimation of the
optimal learning rate. Other examples of adaptive learning rate strategies
are discussed below (Sec.~\ref{sec:adaptive}).

\item The {\bf mini-batch size} ($B$ in Eq.~\eqref{eq:minibatch-update}) is typically 
chosen between 1 and a few hundreds, e.g. $B=32$ is a good default value, 
with values above $10$ taking advantage of the speed-up of matrix-matrix products over
matrix-vector products. The impact of $B$ is mostly computational,
i.e., larger $B$ yield faster computation (with appropriate implementations)
but requires visiting more examples in order to reach the same error, since
there are less updates per epoch.
In theory, this hyper-parameter should impact {\em training time}
and not so much {\em test performance}, so it can be optimized separately
of the other hyper-parameters, by comparing training curves (training
and validation error vs amount of training time), after the other
hyper-parameters (except learning rate) have been selected. $B$
and $\epsilon_0$ may slightly interact with other hyper-parameters
so both should be re-optimized at the end. Once $B$ is selected, it
can generally be fixed while the other hyper-parameters can be further
optimized (except for a {\em momentum} hyper-parameter, if one is used).
\item {\bf Number of training iterations} $T$ (measured in mini-batch updates).
This hyper-parameter is particular in that it can be optimized almost for
free using the principle of {\em early stopping}: by keeping track of
the out-of-sample error (as for example estimated on a validation set)
as training progresses (every $N$ updates), one can decide how long to train for any
given setting of all the other hyper-parameters. Early stopping is an inexpensive
way to avoid strong overfitting, i.e., even if the other hyper-parameters
would yield to overfitting, early stopping will considerably reduce
the overfitting damage that would otherwise ensue. It also means that it hides
the overfitting effect of other hyper-parameters, possibly obscuring
the analysis that one may want to do when trying to figure out the
effect of individual hyper-parameters, i.e., it tends to even out
the performance obtained by many otherwise overfitting configurations of 
hyper-parameters by compensating a too large capacity with a smaller
training time. For this reason, it might be useful to turn early-stopping
off when analyzing the effect of
individual hyper-parameters. Now let us turn to implementation details. Practically, one needs
to continue training beyond the selected number of training iterations $\hat{T}$ (which should
be the point of lowest validation error in the training run) in order to ascertain that validation error 
is unlikely to go lower than at the selected point. A heuristic
introduced in the {\em Deep Learning Tutorials}\footnote{~\tt http://deeplearning.net/tutorial/}
is based on the idea of {\em patience} (set initially to 10000 examples in the MLP tutorial),
which is a minimum number of training examples to see after the candidate
selected point $\hat{T}$ before deciding to stop training (i.e. before accepting 
this candidate as the
final answer). As training proceeds and new candidate selected
points $\hat{T}$ (new minima of the validation error) are observed, the patience parameter
is increased,
either multiplicatively or additively on top of the last $\hat{T}$ found. Hence,
if we find a new minimum\footnote{~Ideally, we should use a statistical test of significance
and accept a new minimum (over a longer training period) only if the improvement is statistically
significant, based on the size and variance estimates one can compute for the validation set.}
 at $t$, we save the
current best model, update
$\hat{T}\leftarrow t$ and we increase our patience up to $t+$constant or $t\times$ constant.
Note that validation error should not be estimated after each
training update (that would be really wasteful) but after every $N$ examples, where
$N$ is at least as large as the validation set (ideally several times larger so that
the early stopping overhead remains small)\footnote{~When an extra processor on the same
machine is available, validation error can conveniently be recomputed by a processor
different from the one performing the training updates, allowing more frequent 
computation of validation error.}.
\item {\bf Momentum} $\beta$. It has long been advocated~\citep{Hinton78,Hinton-RBMguide}
to temporally smooth out the stochastic gradient samples obtained during
the stochastic gradient descent. For example, a moving average of the past gradients
can be computed with $\bar{g}\leftarrow(1-\beta)\bar{g}+\beta g$,
where $g$ is the instantaneous gradient $\frac{\partial L(z_t,\theta)}{\partial \theta}$ 
or a minibatch average, and
$\beta$ is a small positive coefficient that controls how fast the old examples get downweighted
in the moving average.
The simplest momentum trick is to make
the updates proportional to this smoothed gradient estimator $\bar{g}$
instead of the instantaneous gradient $g$.
The idea is that it removes some of the noise and oscillations that
gradient descent has, in particular in the directions of high curvature of the
loss function\footnote{~Think about a ball coming down a valley.  Since it has not started from the
bottom of the valley it will oscillate between its sides as it settles deeper, forcing
the learning rate to be small to avoid large oscillations that would kick it
out of the valley. Averaging out the local gradients along the way will cancel the opposing
forces from each side of the valley.}. A default value of $\beta=1$ (no momentum)
works well in many cases but in some cases momentum seems to make a positive difference.
Polyak averaging~\citep{Polyak+Juditsky-1992} is a related form of parameter averaging\footnote{~Polyak averaging 
uses for predictions a moving average of the parameters found in the trajectory of stochastic gradient descent.}
that has theoretical advantages and has been advocated and shown to bring improvements
on some unsupervised learning procedures such as RBMs~\citep{Swersky+al-2010}.
More recently, several mathematically motivated algorithms~\citep{Nesterov-2009,LeRoux-arxiv-2012} 
have been proposed that
incorporate some form of momentum and that also ensure much faster convergence 
(linear rather than sublinear) 
compared to stochastic gradient descent, at least for convex optimization problems.
See also~\cite{Bottou-tricks-chapter-2013} for an example of averaged SGD with successful
empirical speedups in the convex case.
Note however that in the pure online case (stream of examples) and under some assumptions, 
the sublinear rate
of convergence of stochastic gradient descent with $O(1/t)$ decrease
of learning rate is an optimal rate, at least for convex problems~\citep{Nemirovski+Yudin-83}.
That would suggest that for really large training sets it may not be
possible to obtain better rates than ordinary stochastic gradient descent,
albeit the constants in front (which depend on the condition number of
the Hessian) may still be greatly reduced by using second-order information
online~\citep{bottou-lecun-04b-small,bottou-bousquet-2008-small}.
\item {\bf Layer-specific optimization hyper-parameters}: although rarely done, it is
possible to use different values of optimization hyper-parameters (such as
the learning rate) on different layers of a multi-layer network. This is especially
appropriate (and easier to do) in the context of layer-wise unsupervised pre-training,
since each layer is trained separately (while the layers below are kept fixed).
This would be particularly useful when the number of units per layer varies
a lot from layer to layer. See the paragraph below entitled {\bf Layer-wise optimization
of hyper-parameters} (Sec.~\ref{sec:layer-wise}). Some researchers also advocate
the use of different learning rates for the different {\em types} of parameters
one finds in the model, such as biases and weights in the standard multi-layer network,
but the issue becomes more important when parameters such as precision or variance
are included in the lot~\citep{Courville+al-2011-small}.
\end{itemize}
Up to now we have only discussed the hyper-parameters in the setup where
one trains a neural network by stochastic gradient descent. With other
optimization algorithms, some hyper-parameters are typically different. For
example, Conjugate Gradient (CG) algorithms typically have a number of
line search steps (which is a hyper-parameter) and a tolerance for stopping
each line search (another hyper-parameter). An optimization algorithm like
L-BFGS (limited-memory Broyden–Fletcher–Goldfarb–Shanno)
also has a hyper-parameter controlling the
memory usage of the algorithm, the rank of the Hessian approximation kept
in memory, which also has an influence on the efficiency of each step. Both
CG and L-BFGS are iterative (e.g., one line search per iteration), and the
number of iterations can be optimized as described above for stochastic
gradient descent, with early stopping.

\subsection{Hyper-Parameters of the Model and Training Criterion}

Let us now turn to ``model'' and ``criterion'' hyper-parameters typically found in neural networks,
especially deep neural networks.
\begin{itemize}
\item {\bf Number of hidden units $n_h$}. Each layer in a multi-layer neural network
typically has a size that we are free to set and that controls capacity.
Because of early stopping and possibly other regularizers (e.g., weight decay, 
discussed below),
it is mostly important to choose $n_h$ large enough. Larger than optimal
values typically do not hurt generalization performance much, but of course they
require proportionally more computation (in $O(n_h^2)$ if scaling all
the layers at the same time in a fully connected architecture). Like
for many other hyper-parameters, there is the option of allowing a
different value of $n_h$ for each hidden layer\footnote{~A hidden layer is
a group of units that is neither an input layer
nor an output layer.} of a deep architecture. See the paragraph below entitled {\bf Layer-wise optimization
of hyper-parameters} (Sec.~\ref{sec:layer-wise}).
In a large comparative study~\citep{Larochelle-jmlr-2009-small}, we found that
using the same size for all layers worked generally better or the same as
using a decreasing size (pyramid-like) or increasing size (upside down pyramid),
but of course this may be data-dependent. For most tasks that we worked on, we find that
an {\em overcomplete}\footnote{~larger than the input vector} 
first hidden layer works better than an undercomplete one.
Another even more often
validated empirical observation is that the optimal $n_h$ is much larger
when using unsupervised pre-training in a supervised neural network, e.g.,
going from hundreds of units to thousands of units. A plausible explanation
is that after unsupervised pre-training many of the hidden units are carrying information
that is irrelevant to the specific supervised task of interest. In order to
make sure that the information relevant to the task is captured, larger hidden
layers are therefore necessary when using unsupervised pre-training.
\item {\bf Weight decay} regularization coefficient $\lambda$. A way
to reduce overfitting is to add a {\em regularization term} to the
training criterion, which limits the capacity of the learner. The parameters
of machine learning models can be regularized by pushing them towards
a prior value, which is typically 0. L2 regularization adds a term $\lambda \sum_i \theta_i^2$
to the training criterion, while L1 regularization adds a term $\lambda \sum_i |\theta_i|$.
Both types of terms can be included. There is a clean Bayesian
justification for such a regularization term: it is the negative log-prior $-\log P(\theta)$
on the parameters $\theta$. The training criterion then corresponds
to the negative joint likelihood of data and parameters, $-\log P(data,\theta) = -\log P(data|\theta) -\log P(\theta)$,
with the loss function $L(z,\theta)$ being interpreted as $-\log P(z|\theta)$ and
$-\log P(data|\theta)=-\sum_{t=1}^T L(z_t,\theta)$ if the $data$ consists of
$T$ i.i.d. examples $z_t$. This detail is important to note because when one
is doing stochastic gradient-based learning, it makes sense to use an
unbiased estimator of the gradient of the total training criterion (including both the
total loss and the regularizer), but one only considers a single mini-batch or example at a time. How
should the regularizer be weighted in this sum, which is different from the sum of the
regularizer and the total loss on all examples? 
On each mini-batch update, the gradient of the regularization penalty should be multiplied
not just by $\lambda$ but also by $\frac{B}{T}$, i.e., one over the number of 
updates needed to go once through the training set. When the training set size
is not a multiple of $B$, the last mini-batch will have size $B'<B$ and the
contribution of the regularizer to the mini-batch gradient should therefore be modified 
accordingly (i.e. scaled by $\frac{B'}{B}$ compared to other mini-batches).
In the pure online setting (there
is no fixed ahead training set size nor iterating again on the examples),
it would then make sense to use $\frac{B}{t}$ at example $t$, or one over the
number of updates to date.
L2 regularization penalizes large
values more strongly and corresponds to a Gaussian prior $\propto \exp(-\frac{1}{2} \frac{||\theta||^2}{\sigma^2})$ 
with prior variance $\sigma^2=1/(2 \lambda)$.
Note that there is a connection between early stopping (see above, choosing
the number of training iterations) and L2 regularization~\citep{Collobert2004-small},
with one basically playing the same role as the other (but early stopping
allowing a much more efficient selection of the hyper-parameter value, which
suggests dropping L2 regularization altogether when early-stopping is used).
However, L1 regularization behaves differently and can sometimes be useful,
acting as a form of feature selection.
L1 regularization makes sure that parameters that are not
really very useful are driven to zero (i.e. encouraging sparsity of the
parameter values), and corresponds to a Laplace density 
prior $\propto e^{-\frac{|\theta|}{s}}$ with scale parameter $s=\frac{1}{\lambda}$.
L1 regularization often helps to make the input filters\footnote{~The input weights
of a 1st layer neuron are often called ``filters'' because of analogies with
signal processing techniques such as convolutions.} cleaner (more spatially localized)
and easier to interpret.
Stochastic gradient descent will not yield actual zeros but values hovering around zero.
If both L1 and L2 regularization are used, a different coefficient 
(i.e. a different hyper-parameter) should be considered for each, and one may also use a different
coefficient for different layers. In particular, the input weights and output
weights may be treated differently. 

One reason for treating output weights
differently (i.e., not relying only on early stopping)
is that we know that it is sufficient to regularize only the
output weights in order to constrain capacity: in the limit case of
the number of hidden units going to infinity, L2 regularization corresponds
to Support Vector Machines (SVM) while L1 regularization corresponds to boosting~\citep{Bengio+al-2005-small}.
Another reason for treating inputs and outputs differently from hidden units is because
they may be sparse. For example, some input features may be 0 most of the time
while others are non-zero frequently. In that case, there are fewer examples
that inform the model about that rarely active input feature, and the
corresponding parameters (weights outgoing from the corresponding input units)
should be more regularized than the parameters associated with frequently
observed inputs. A similar situation may occur with target variables that
are sparse (e.g., trying to predict rarely observed events). In both cases,
the effective number of meaningful updates seen by these parameters is less than
the actual number of updates. This suggests to scale the regularization coefficient
of these parameters by one over the effective number of updates seen by the
parameter. A related formula turns up in Bayesian probit
regression applied to sparse inputs~\citep{Graepel-2010-small}.
Some practitioners also choose to penalize
only the weights $w$ and not the biases $b$ associated with the hidden unit
activations $w'z+b$ for a unit taking the vector of values $z$ as input.
This guarantees that even with strong regularization, the predictor would
converge to the optimal constant predictor, rather than the one corresponding
to 0 activation. For example, with the mean-square loss and the cross-entropy loss, the optimal constant
predictor is the output average.
\item {\bf Sparsity of activation} regularization coefficient $\alpha$.  A
  common practice in the Deep Learning
  literature~\citep{ranzato-07-small,ranzato-08-small,HonglakL2008-small,HonglakL2009-small,Bradley+Bagnell-2009-small,Glorot+al-AI-2011-small,Coates2011b,goodfellow+all-NIPS2011}
  consists in adding a penalty term to the training criterion that
  encourages the hidden units to be sparse, i.e., with values at or near 0.
  Although the L1 penalty (discussed above in the case of weights) 
  can also be applied to hidden units
  activations, this is mathematically very different from the L1
  regularization term on parameters. Whereas the latter corresponds to a
  prior on the parameters, the former does not because it involves the
  training distribution (since we are looking at data-dependent hidden units outputs).
  Although we will not discuss this much here, the inspiration for
  a sparse representation in Deep Learning comes from the earlier work on
  {\em sparse coding}~\citep{Olshausen-97-small}. As discussed
  in~\citet{Goodfellow2009-small} sparse representations may be advantageous
  because they encourage representations that {\em disentangle} the
  underlying factors of representation. A sparsity-inducing penalty is also
  a way to regularize (in the sense of reducing the number of examples that
  the learner can learn by heart)~\citep{ranzato-08-small}, which means that the
  sparsity coefficient is likely to interact with the many other
  hyper-parameters which influence capacity.  In general, increased
  sparsity can be compensated by a larger number of hidden units.

Several approaches have been proposed to induce a sparse representation (or with more hidden units
whose activation is closer to 0). One approach~\citep{ranzato-08-small,Le-ICML2011-small,Zou-Ng-Yu-NIPSwkshop2011} 
is simply to penalize the L1 norm of the representation 
or another function of the hidden units' activation (such as the student-t log-prior).
This typically makes sense for non-linearities such as the sigmoid which have a saturating
output around 0, but not for  the hyperbolic tangent non-linearity (whose saturation is near
the -1 and 1 interval borders rather than near the origin). Another option is to
penalize the biases of the hidden units, to make them more 
negative~\citep{ranzato-07-small,HonglakL2008-small,Goodfellow2009-small,Larochelle+Bengio-2008-small}.
Note that penalizing the 
bias runs the danger that the weights could compensate for the bias\footnote{~because the
input to the layer generally has a non-zero average, that when multiplied by the weights
acts like a bias},
which could hurt the numerical optimization of parameters.
When directly penalizing the hidden unit outputs, several variants
can be found in the literature, but no clear comparative analysis
has been published to evaluate which one works better. Although the L1
penalty (i.e., simply $\alpha$ times the sum of output elements $h_j$ in the case of sigmoid non-linearity)
would seem the most natural (because of its use in sparse coding), it
is used in few papers involving sparse auto-encoders. A close cousin of
the L1 penalty is the Student-t penalty ($\log(1+h_j^2)$), originally proposed for
sparse coding~\citep{Olshausen-97-small}. Several researchers
penalize the {\em average} output $\bar{h}_j$ (e.g. over a mini-batch), and instead
of pushing it to 0, encourage it to approach a fixed target $\rho$. This can be done through
a mean-square error penalty such as $\sum_j (\rho-\bar{h}_j)^2$, or 
maybe more sensibly (because $h_j$ behaves like a probability), 
a Kullback-Liebler divergence  with respect to the binomial distribution with
probability $\rho$, $-\rho \log \bar{h}_j -(1-\rho)\log(1-\bar{h}_j)+$constant, 
e.g., with $\rho=0.05$, as in~\citep{Hinton-RBMguide}.
In addition to the regularization penalty itself, the choice of activation function
can have a strong impact on the sparsity obtained. In particular, rectifying non-linearities
(such as $\max(0,x)$, instead of a sigmoid) have been very successful in several 
instances~\citep{Jarrett-ICCV2009-small,Nair+Hinton-2010-small,Glorot+al-AI-2011-small,UTLC+LISA-2011-small,Glorot+al-ICML-2011-small}. The rectifier also relates to the
hard tanh~\citep{Collobert+SBengio-2004}, whose derivatives are also 0 or 1.
In sparse coding and sparse predictive coding~\citep{Koray-08-small}
the activations are directly optimized and actual zeros are the expected result of the optimization.
In that case, ordinary stochastic gradient is not guaranteed to find these zeros (it will
oscillate around) and other methods such as proximal gradient are more appropriate~\citep{Bertsekas-2010}.
\item {\bf Neuron non-linearity}. The typical neuron output is $s(a)=s(w'x + b)$, where
$x$ is the vector of inputs into the neuron, $w$ the vector of weights and $b$ the
offset or bias parameter, while $s$ is a scalar non-linear function.
Several non-linearities have been proposed and some choices of
non-linearities have been
shown to be more successful~\citep{Jarrett-ICCV2009-small,GlorotAISTATS2010-small,Glorot+al-AI-2011-small}.
The most commonly used by the author, for hidden units,
are the sigmoid $1/(1+e^{-a})$, the hyperbolic
tangent $\frac{e^a-e^{-a}}{e^a+e^{-a}}$, the rectifier $\max(0,a)$
and the hard tanh~\citep{Collobert+SBengio-2004}.
Note that the sigmoid was shown to yield serious optimization
difficulties when used as the top hidden layer of a deep supervised network
~\citep{GlorotAISTATS2010-small} without unsupervised pre-training, but works
well for auto-encoder variants\footnote{~The author hypothesizes that this
discrepency is due to the fact that the weight matrix $W$ of an auto-encoder
of the form $r(x)=W^T {\rm sigmoid}(W x)$ is pulled towards being orthonormal since this would make the auto-encoder 
closer to the identity function, because $W^T W x \approx x$ when $W$ is orthonormal
and $x$ is in the span of the rows of $W$.}.
For output (or reconstruction) units, hard neuron non-linearities like the rectifier
do not make sense because when the unit is saturated (e.g. $a<0$ for the
rectifier) and associated with a loss, no gradient is propagated inside
the network, i.e., there is no chance to correct the error\footnote{~A hard
non-linearity for the output units non-linearity is very different from a
hard non-linearity in the loss function, such as the hinge loss. In the
latter case the derivative is 0 only when there is no error.}.
In the case of hidden layers the gradient manages to go through
a subset of the hidden units, even if the others are saturated.
For output units a good trick is to obtain the output non-linearity and the loss by
considering the associated negative log-likelihood and choosing an appropriate
(conditional) output probability model, usually in the exponential family.
For example, one can typically take squared error and linear outputs to correspond to
a Gaussian output model, cross-entropy and sigmoids to correspond to a binomial
output model, and $-\log {\rm output}[\mbox{target class}]$ with softmax outputs
to correspond to multinomial output variables. For reasons yet to be elucidated,
having a sigmoidal non-linearity on the output (reconstruction) units (along with
target inputs normalized in the (0,1) interval) seems to
be helpful when training the contractive auto-encoder.
\item {\bf Weights initialization scaling coefficient}. Biases
can generally be initialized to zero but weights need to be initialized
carefully to break the symmetry between hidden units of the same 
layer\footnote{~By symmetry, if hidden units of the same layer share the
same input and output weights, they will compute the same output and
receive the same gradient, hence performing the same update and remaining
identical, thus wasting capacity.}.
Because different output units receive different gradient signals,
this symmetry breaking issue does not concern the output weights (into the output units), which can therefore
also be set to zero.
Although several tricks~\citep{LeCun+98backprop-small,GlorotAISTATS2010-small} 
for initializing the weights into hidden layers have been proposed (i.e. a hyper-parameter is the discrete
choice between them),~\citet{Bergstra+Bengio-2012-small}
also inserted as an extra hyper-parameter a scaling coefficient for the initialization
range. These tricks are
based on the idea that units with more inputs (the {\em fan-in} of the unit)
should have smaller weights.
Both~\citet{LeCun+98backprop-small} and \citet{GlorotAISTATS2010-small} recommend
scaling by the inverse of the {\em square root} of the fan-in, although
\citet{GlorotAISTATS2010-small} and the Deep Learning Tutorials use a combination
of the fan-in and fan-out, e.g.,  sample a Uniform($-r,r)$
with $r=\sqrt{6/(\mbox{fan-in}+\mbox{fan-out})}$ for hyperbolic tangent units
and $r=4\sqrt{6/(\mbox{fan-in}+\mbox{fan-out})}$ for sigmoid units.
We have found that we could avoid any hyper-parameter related to initialization
using these formulas (and the derivation in \citet{GlorotAISTATS2010-small} can be used
to derive the formula for other settings).  Note however that in the case of
RBMs, a zero-mean Gaussian with a small standard deviation around 0.1 or 0.01
works well~\citep{Hinton-RBMguide} to initialize the weights, while visible
biases are typically set to their optimal value if the weights were 0, i.e.,
$\log(\bar{x}/(1-\bar{x}))$ in the case of a binomial visible unit whose corresponding
binary input feature has empirical mean $\bar{x}$ in the training set.

An important choice is whether one should use {\em unsupervised pre-training}
(and which unsupervised feature learning algorithm to use) in order to initialize
parameters. In most settings we have found unsupervised pre-training to help
and very rarely to hurt, but of course that implies additional training time
and additional hyper-parameters.
\item {\bf Random seeds.} There are often several sources of randomness
in the training of neural networks and deep learners (such as for random
initialization, sampling examples, sampling hidden units in stochastic models
such as RBMs, or sampling corruption noise in denoising auto-encoders).
Some random seeds could therefore yield better results than others.
Because of the presence of local minima in the training criterion of
neural networks (except in the linear case or with fixed lower layers),
parameter initialization matters.
See~\citet{Erhan+al-2010-small} for an example of histograms of
test errors for hundreds of different random seeds. Typically, the
choice of random seed only has a slight effect on the result and can
mostly be ignored in general or for most of the hyper-parameter search process. 
If computing power is available, then a final set of jobs  with different random seeds (5 to 10)
for a small set of best choices of hyper-parameter values can
squeeze a bit more performance. Another way to exploit computing power
to push performance a bit is model averaging, as in Bagging~\citep{ML:Breiman:bagging}
and Bayesian methods.
After training them,
the outputs of different networks
(or in general different learning algorithms) can be averaged. For example, the difference between the
neural networks being averaged into a committee 
may come from the different seeds used for parameter initialization, or the use of
different subsets of input variables, or different subsets of
training examples (the latter being called Bagging).
\item {\bf Preprocessing.} Many preprocessing steps have been proposed to
  massage raw data into appropriate inputs for neural networks and model
  selection must also choose among them. In addition to element-wise
  standardization (subtract mean and divide by standard deviation),
  Principal Components Analysis (PCA) has often been
  advocated~\citep{LeCun+98backprop-small,Bergstra+Bengio-2012-small} and also allows
  dimensionality reduction, at the price of an extra hyper-parameter (the
  number of principal components retained, or the proportion of variance
  explained). A convenient non-linear preprocessing is the {\em
    uniformization}~\citep{UTLC+LISA-2011-small} of each feature (which
  estimates its cumulative distribution $F_i$ and then transforms each
  feature $x_i$ by its quantile $F_i^{-1}(x_i)$, i.e., returns an
  approximate normalized rank or quantile for the value $x_i$). A simpler to compute
  transform that may help reduce the tails of input features is a non-linearity such as the logarithm
  or the square root, in an attempt to make them more Gaussian-like.
\end{itemize}

In addition to the above somewhat generic choices, more choices arise with
different architectures and learning algorithms. For example, the denoising
auto-encoder has a hyper-parameter scaling the amount of input
corruption and the contractive auto-encoder has as hyper-parameter
a coefficient scaling the norm of the Jacobian of the encoder, i.e.,
controlling the importance of the contraction penalty. The latter seems
to be a rather sensitive hyper-parameter that must be tuned
carefully. The contractive auto-encoder's success also seems
sensitive to the {\bf weight tying} constraint used in many
auto-encoder architectures: the decoder's weight matrix
is equal to the transpose of the encoder's weight matrix.
The specific architecture used in the contractive auto-encoder
(with tied weights, sigmoid non-linearies on hidden and
reconstruction units, along with squared loss or cross-entropy
loss) works quite well but other related variants do not
always train well, for reasons that remain to be understood.

There are also many architectural choices that are relevant
in the case of convolutional architectures (e.g. for modeling images,
time-series or sound)~\citep{LeCun89,LeCun98-small,Le2010-small} in which hidden units have
local receptive fields. Their discussion is postponed to
another chapter~\citep{LeCun-tricks-chapter}.

\subsection{Manual Search and Grid Search}

Many of the hyper-parameters or model choices described above can be ignored
by picking a standard trick suggested here or in some other paper. Still, one
remains with a substantial number of choices to be made, which may give the
impression of neural network training as an art. With modern computing
facilities based on large computer clusters, it is however possible
to make the optimization of hyper-parameters a more reproducible and automated process,
using techniques such as grid search or better, random search, or
even hyper-parameter optimization, discussed below.

\subsubsection{General guidance for the exploration of hyper-parameters}

First of all, let us consider recommendations for exploring hyper-parameter
settings, whether with manual search, with an automated procedure, or with a combination
of both. We call a {\em numerical hyper-parameter} one that involves choosing
a real number or an integer (where order matters), 
as opposed to making a discrete symbolic choice from an unordered set.
Examples of numerical hyper-parameters are regularization coefficients, number
of hidden units, number of training iterations, etc. One has to think of
hyper-parameter selection as {\em a difficult form of learning}: there is both
an optimization problem (looking for hyper-parameter configurations that yield
low validation error) and a generalization problem: there is uncertainty about the
expected generalization after optimizing validation performance, and it is possible to overfit
the validation error and get optimistically biased estimators of performance
when comparing many hyper-parameter configurations. The training
criterion for this learning is typically the validation set error, which is a proxy for
generalization error. Unfortunately, the relation between hyper-parameters
and validation error can be complicated. Although to first approximation
we expect a kind of U-shaped curve (when considering only a single
hyper-parameter, the others being fixed), this curve can also have noisy
variations, in part due to the use of finite data sets.

\begin{itemize}
\item {\bf Best value on the border.} When considering the validation error obtained for different values of a 
numerical hyper-parameter one should pay attention as to whether or not the best value found is near the border
of the investigated interval. If it is near the border, then this suggests that better values can
be found with values beyond the border: it is recommended in that case
to explore further, beyond that border.
Because the relation between a hyper-parameter and validation error can be
noisy, it is generally not enough to try very few values. For instance,
trying only 3 values for a numerical hyper-parameter is insufficient,
even if the best value found is the middle one.
\item {\bf Scale of values considered.} Exploring values of a numerical hyper-parameter
entails choosing a {\em starting interval} to be searched, which is therefore a kind of hyper-hyper-parameter.
By choosing the interval large enough to start with, but based on previous experience
with this hyper-parameter, we ensure that we do not get completely wrong results.
Now instead of choosing the intermediate values {\em linearly} in the chosen interval,
it often makes much more sense to consider a linear or uniform sampling in the {\em log-domain}
(in the space of the logarithm of the hyper-parameter). For example,
the results obtained with a learning rate of 0.01 are likely to be very
similar to the results with 0.011 while results with 0.001 could be quite
different from results with 0.002 even though the absolute difference is the same in
both cases. The {\em ratio} between different values is often a better guide of the
expected impact of the change. That is why exploring uniformly or regularly-spaced values
in the space of the {\em logarithm} of the numerical
hyper-parameter is typically preferred for positive-valued numerical hyper-parameters.
\item {\bf Computational considerations.} Validation error is actually not
the only measure to consider in selecting hyper-parameters. Often, one has
to consider computational cost, either of training or prediction. Computing
resources for training and prediction are limited and generally condition
the choice of intervals of considered values: for example increasing the
number of hidden units or number of training iterations also scales up computation.
An interesting idea is to use {\em computationally cheap estimators} of
validation error to select some hyper-parameters. For example,~\citet{Saxe-ICML2011-small}
showed that the architecture hyper-parameters of convolutional networks could
be selected using {\em random weights} in the lower layers of the network
(filters of the convolution).
While this yields a noisy and biased (pessimistic) estimator of the validation error
which would otherwise be obtained with full training, this cheap estimator appears 
to be correlated with the expensive validation error. Hence this cheap estimator is enough
for selecting some hyper-parameters (or for keeping under consideration for further and more expensive
evaluation only the few best choices found). Even without cheap estimators
of generalization error, high-throughput computing (e.g., on clusters, GPUs, or clusters
of GPUs) can be exploited to run not just hundreds but thousands of training jobs,
something not conceivable only a few years ago, with each job taking on the order of
hours or days for larger datasets. With computationally cheap surrogates, some researchers have
run on the order of ten thousands trials, and we can expect future advances in parallelized
computing power to boost these numbers.
\end{itemize}

\subsubsection{Coordinate Descent and Multi-Resolution Search}

When performing a manual search and with access to only a single computer, a reasonable
strategy is {\em coordinate descent}: change only one hyper-parameter at a time, always
making a change from the best configuration of hyper-parameters found up to now. Instead
of a standard coordinate descent (which systematically cycles through all the variables
to be optimized) one can make sure to regularly fine-tune the most sensitive variables, such
as the learning rate.

Another important idea is that there is no point in exploring the effect of fine changes
before one or more reasonably good settings have been found. The idea of {\em multi-resolution
search} is to start the search by considering only a few values of the numerical hyper-parameters
(over a large range), or considering large changes each time a new value is tried.
One can then start from the one or few best configurations found and explore more
locally around them with smaller variations around these values. 

\subsubsection{Automated and Semi-automated Grid Search}

Once some interval or set of values has been selected for each hyper-parameter
(thus defining a search space), a simple strategy that exploits parallel computing
is the {\bf grid search}. One first needs to convert the numerical intervals into
lists of values (e.g., $K$ regularly-spaced values in the log-domain of the hyper-parameter).
The grid search is simply an exhaustive search through all the combinations of these
values. The cross-product of these lists contains a number of elements that is
unfortunately exponential in the number of hyper-parameters (e.g., with 5 hyper-parameters,
each allowed to take 6 different values, one gets $6^5=7776$ configurations). In
section~\ref{sec:random-sampling} below we consider an approach that works
more efficiently than the grid search when the number of hyper-parameters 
increases beyond 2 or 3.

The advantage of the grid search, compared to many other optimization strategies (such as
coordinate descent), is that it is fully parallelizable. If a large computer cluster
is available, it is tempting to choose a model selection strategy that can take
advantage of parallelization. One practical disadvantage of grid search (especially
against random search, Sec.~\ref{sec:random-sampling}), with a parallelized
set of jobs on a cluster, is that if only one of the jobs fails\footnote{~For all kinds
of hardware and software reasons, a job failing is very common.} then one has to launch
another volley of jobs to complete the grid (and yet a third one if any of these
fails, etc.), thus multiplying the overall computing time.

Typically, a single grid search is not enough and practitioners tend
to proceed with a sequence of grid searches, each time adjusting the
ranges of values considered based on the previous results obtained.
Although this can be done manually, this procedure can also be automated
by considering the idea of multi-resolution search to guide this outer loop.
Different, more local, grid searches can be launched in the neighborhood
of the best solutions found previously. In addition, the idea of
coordinate descent can also be thrown in, by making each grid search
focus on only a few of the hyper-parameters.  For example, it is common
practice to start by exploring the initial learning rate while keeping fixed (and
initially constant) the learning rate descent schedule. Once the shape
of the schedule has been chosen, it may be possible to further refine the
learning rate, but in a smaller interval around the best value found.

Humans can get very good at performing hyper-parameter search, and having
a human in the loop also has the advantage that it can help detect bugs
or unwanted or unexpected behavior of a learning algorithm. However, for
the sake of reproducibility, machine learning researchers should strive
to use procedures that do not involve human decisions in the middle,
only at the outset (e.g., setting hyper-parameter ranges, which can
be specified in a paper describing the experiments).

\subsubsection{Layer-wise optimization of hyper-parameters}
\label{sec:layer-wise}

In the case of Deep Learning with unsupervised pre-training there is an
opportunity for combining coordinate descent and cheap relative validation
set performance evaluation associated with some hyper-parameter
choices. The idea, described by~\citet{UTLC+LISA-2011-small,UTLC+DL+tutorial-2011-small}, is to perform
greedy choices for the hyper-parameters associated with lower layers (near
the input) before training the higher layers.  One first trains
(unsupervised) the first layer with different hyper-parameter values and
somehow estimates the relative validation error that would be obtained from
these different configurations if the final network only had this single
layer as internal representation.  In the common case where the ultimate
task is supervised, it means training a simple supervised predictor (e.g. a
linear classifier) on top of the learned representation. In the case of
a linear predictor (e.g. regression or logistic regression) this can even be
done on the fly while unsupervised training of the representation
progresses (i.e. can be used for early stopping as well), as
in~\citep{Larochelle-jmlr-2009-small}. Once a set of apparently good
(according to this greedy evaluation) hyper-parameters values
has been found (or possibly using only the best one found), these good
values can be used as starting point to train (and hyper-optimize)
a second layer in the same way, etc. The completely greedy approach
is to keep only the best configuration up to now (for the lower layers),
but keeping the $K$ best configurations overall only multiplies computational
costs of hyper-parameter selection by $K$ for layers beyond the first one,
because we would still keep only the best $K$ configurations from all
the 1st layer and 2nd layer hyper-parameters as starting points for
exploring 3rd layer hyper-parameters, etc. This procedure is formalized
in the Algorithm~\ref{algo:greedy-layerwise-hyperopt} below.
\begin{algorithm}
\caption{{\bf: Greedy layer-wise hyper-parameter optimization.}}
\label{algo:greedy-layerwise-hyperopt}
\begin{algorithmic}
\STATE {\bf input} $K$: number of best configurations to keep at each level.
\STATE {\bf input} $NLEVELS$: number of levels of the deep architecture
\STATE {\bf input} $LEVELSETTINGS$: list of hyper-parameter settings to be considered for unsupervised pre-training of a level
\STATE {\bf input} $SFTSETTINGS$: list of hyper-parameter settings to be considered for supervised fine-tuning
\STATE
\STATE Initialize set of best configurations $S=\emptyset$
\FOR{$L=1$ {\bf to} $NLEVELS$}
  \FOR{$C$ {\bf in} $LEVELSETTINGS$}
    \FOR{$H$ {\bf in} ($S$ or $\{\emptyset\}$)}
      \STATE * Pretrain level $L$ using hyper-parameter setting $C$ for level $L$ and 
      the parameters obtained with setting $H$ for lower levels.
      \STATE * Evaluate target task performance $\cal L$ using this depth-$L$ pre-trained architecture
      (e.g. train a linear classifier on top of these layers and estimate validation error).
      \STATE * Push the pair $(C \cup H,{\cal L})$ into $S$ if it is among the $K$ best performing of $S$.
    \ENDFOR
  \ENDFOR
\ENDFOR
\FOR{$C$ {\bf in} $SFTSETTINGS$}
  \FOR{$H$ {\bf in} $S$}
    \STATE * Supervised fine-tuning of the pre-trained architecture associated with $H$,
    using supervised fine-tuning hyper-parameter setting $C$.
    \STATE * Evaluate target task performance $\cal L$ of this fine-tuned predictor (e.g. validation error).
    \STATE * Push the pair $(C \cup H,{\cal L})$ into $S$ if it is among the $K$ best performing of $S$.
  \ENDFOR
\ENDFOR
\STATE {\bf output} $S$ the set of $K$ best-performing models with their settings and validation performance.
\end{algorithmic}
\end{algorithm}
Since greedy layer-wise pre-training does not modify the lower layers
when pre-training the upper layers, this is also very efficient
computationally. This procedure allows one to set the hyper-parameters
associated with the unsupervised pre-training stage, and then
there remains hyper-parameters to be selected for the supervised
fine-tuning stage, if one is desired. A final supervised fine-tuning
stage is strongly suggested, especially when there are many
labeled examples~\citep{Lamblin-NIPS2010-workshop}.

\subsection{Random Sampling of Hyper-Parameters}
\label{sec:random-sampling}

A serious problem with the grid search approach to find good
hyper-parameter configurations is that it scales
exponentially badly with the number of hyper-parameters considered. In the
above sections we have discussed numerous hyper-parameters and if all of
them were to be explored at the same time it would be impossible to use
only a grid search to do so. 

One may think that there are no other options simply because this is an instance
of the curse of dimensionality. But like we have found in our work
on Deep Learning~\citep{Bengio-2009-book}, if there is some structure in a
target function we are trying to discover, then there is a chance
to find good solutions without paying an exponential price.
It turns out that in many practical cases we have encountered,
there is a kind of structure that {\em random sampling} can
exploit~\citep{Bergstra+Bengio-2012-small}. The idea of random sampling
is to replace the regular grid by a random (typically uniform)
sampling. Each tested hyper-parameter configuration is selected
by independently sampling each hyper-parameter from a prior
distribution (typically uniform in the log-domain, inside the
interval of interest). For a discrete hyper-parameter, a multinomial
distribution can be defined according to our prior beliefs on the likely
good values. At worse, i.e., with no prior preference at all, this would
be a uniform distribution across the allowed values.
In fact, we can use our prior knowledge to make this prior distribution
quite sophisticated. For example, we can readily include knowledge
that some values of some hyper-parameters only make sense in the
context of other particular values of hyper-parameters. This is 
a practical consideration for example when considering layer-specific
hyper-parameters when the number of layers itself is a hyper-parameter.

The experiments performed~\citep{Bergstra+Bengio-2012-small} show that
random sampling can be many times more efficient than grid search as soon
as the number of hyper-parameters goes beyond the 2 or 3 typically seen
with SVMs and vanilla neural networks. The main reason why faster
convergence is observed is because it allows one to explore more values for
each hyper-parameter, whereas in grid search, the same value of a
hyper-parameter is repeated in exponentially many configurations (of all
the other hyper-parameters).  In particular, if only a small subset of the
hyper-parameters really matters, then this procedure can be shown to be
exponentially more efficient. What we found is that for different datasets
and architectures, the subset of hyper-parameters that mattered most was
different, but it was often the case that a few hyper-parameters made a
big difference (and the learning rate is always one of them!). When
marginalizing (by averaging or minimizing) the validation performance to
visualize the effect of one or two hyper-parameters, we get a more noisy
picture using a random search compared to a grid search, because of the random variations of the other hyper-parameters
but one with much more resolution, because so many more different values
have been considered. Practically, one can plot the curves of best validation
error as the number of random trials\footnote{~each random trial
  corresponding to a training job with a particular choice of
  hyper-parameter values} is increased (with mean and standard deviation,
obtained by considering, for each choice of number of trials, all possible
same-size subsets of trials), and this curve tells us that we are
approaching a plateau, i.e., it tells us whether it is worth it or not to
continue launching jobs, i.e., we can perform a kind of early stopping in
the outer optimization over hyper-parameters. Note that one should
distinguish the curve of the ``best trial in first N trials'' with the
curve of the mean (and standard deviation) of the ``best in a subset of
size N''.  The latter is a better statistical representative of the
improvements we should expect if we increase the number of trials. Even if
the former has a plateau, the latter may still be on the increase, pointing
for the need to more hyper-parameter configuration samples, i.e., more
trials~\citep{Bergstra+Bengio-2012-small}.  Comparing these curves with
the equivalent obtained from grid search we see faster convergence with
random search.  On the other hand, note that one advantage of grid search
compared to random sampling is that the qualitative analysis of results is
easier because one can consider variations of a single hyper-parameter with
all the other hyper-parameters being fixed. It may remain a valid option to
do a small grid search around the best solutions found by random search,
considering only the hyper-parameters that were found to matter or which
concern a scientific question of interest\footnote{~This is often the case
  in machine learning research, e.g., does depth of architecture matter?
  then we need to control accurately for the effect of depth, with all
  other hyper-parameters optimized for each value of depth.}.

Random search maintains the advantage of easy parallelization provided by
grid search and improves on it. Indeed, a practical advantage of random
search compared to grid search is that if one of the jobs fails then
there is no need to re-launch that job. It also means that if one has
launched 100 random search jobs, and finds that the convergence curve
still has an interesting slope, one can launch another 50 or 100 without
wasting the first 100. It is not that simple to combine the results
of two grid searches because they are not always compatible (i.e., one
is not a subset of the other).

Finally, although random search is a useful addition to the toolbox
of the practitioner, semi-automatic exploration is still helpful
and one will often iterate between launching a new volley of jobs
and analysis of the results obtained with the previous volley in
order to guide model design and research. What we need is more,
and more efficient, automation of hyper-parameter optimization.
There are some interesting steps in this 
direction~\citep{hutter:2009,Bergstra+al-NIPS2011,hutter+hoos+leyton+brown:2011-small,srinivasan+ramakrishnan:2011}
but much more needs to done.

\section{Debugging and Analysis}
\label{sec:debug}

\subsection{Gradient Checking and Controlled Overfitting}

A very useful debugging step consists in verifying that the implementation
of the gradient $\frac{\partial L}{\partial \theta}$ is compatible with
the computation of $L$ as a function of $\theta$. If the analytically 
computed gradient does not match the one obtained by a finite difference
approximation, this signals that a bug is probably present somewhere.
First of all, looking at for which $i$ one gets important relative 
change between $\frac{\partial L}{\partial \theta_i}$
and its finite difference approximation, we can get hints as to
where the problem may be. An error in sign is particularly troubling, of course.
A good next step is then to verify in the same way 
intermediate gradients $\frac{\partial L}{\partial a}$
with $a$ some quantities that depend on the faulty $\theta$, such as intervening
neuron activations.

As many researchers know, the gradient can be approximated by a finite 
difference approximation obtained from the first-order Taylor expansion of
a scalar function $f$ with respect to a scalar argument $x$:
\[
  \frac{\partial f(x)}{\partial x} = \frac{f(x+\varepsilon)-f(x)}{\varepsilon} + o(\varepsilon)
\]
But a less known fact is that a second order approximation can be achieved
by considering the following alternative formula:
\[
  \frac{\partial f(x)}{\partial x} \approx \frac{f(x+\varepsilon)-f(x-\varepsilon)}{2\varepsilon} 
     + o(\varepsilon^2).
\]
The second order terms of the Taylor expansion of $f(x+\varepsilon)$ and $f(x-\varepsilon)$
cancel each other because they are even, leaving only 3rd or higher order terms,
i.e., $o(\varepsilon^2)$ error after
dividing the difference by $\varepsilon$. Hence this formula is twice more
expensive (not a big deal while debugging) but provides quadratically more precision.

Note that because of finite precision in the computation, there will be a difference
between the analytic (even correct) and finite difference gradient.
Contrary to naive expectations, the relative difference may {\em grow} if
we choose an $\varepsilon$ that is too small, i.e., the error should
first decrease as $\varepsilon$ is decreased, and then may worsen when
numerical precision kicks in, due to non-linearities. We have often used 
a value of $\varepsilon=10^{-4}$ in neural networks, a value that
is sufficiently small to detect most bugs.

Once the gradient is known to be well computed, another sanity check is
that gradient descent (or any other gradient-based optimization)
should be able to overfit on a small training set\footnote{~In principle,
bad local minima could prevent that, but in the overfitting regime, e.g.,
with more hidden units than examples, the global minimum of the training
error can generally be reached almost surely from random initialization,
presumably because the training criterion becomes convex in the parameters
that suffice to get the training error to zero~\citep{Bengio+al-2005-small},
i.e., the output weights of the neural network.}. In particular, to factor
out effects of SGD hyper-parameters, a good sanity check for the code
(and the other hyper-parameters) is to verify that one can overfit on
a small training set using a powerful second order method such as L-BFGS.
For any optimizer, though, as the number of examples
is increased, the degradation of training error should be gradual while
validation error should improve. And one typically sees the advantages
of SGD over batch second-order methods like L-BFGS increase as the
training set size increases. The break-even point may depend on the
task, parallelization (multi-core or GPU, see Sec.\ref{sec:other} below), 
and architecture (number
of computations compared to number of parameters, per example).

Of course, the real goal of learning is to achieve good generalization
error, and the latter can be estimated by measuring performance on an
independent test set. When test error is considered too high, the first
question to ask is whether it is because of a difficulty in optimizing
the training criterion or because of overfitting. Comparing training 
error and test error (and how they change as we change hyper-parameters
that influence capacity, such as the number of training iterations)
helps to answer that question. Depending on the answer, of course, the
appropriate ways to improve test error are different. Optimization difficulties
can be fixed by looking for bugs in the training code, inappropriate values
of optimization hyper-parameters, or simply insufficient capacity
(e.g. not enough degrees of freedom, hidden units, embedding sizes, etc.).
Overfitting difficulties can be addressed by collecting more training
data, introducing more or better regularization terms, multi-task
training, unsupervised pre-training, unsupervised term in the training criterion, 
or considering different function families (or neural network architectures). In
a multi-layer neural network, both problems can be simultaneously present.
For example, as discussed in~\citet{Bengio-nips-2006-small,Bengio-2009-book}, it
is possible to have zero training error with a large top-level hidden
layer that allows the output layer to overfit, while the lower layer
are not doing a good job of extracting useful features because they
were not properly optimized.

Unless using a framework such as Theano which automatically handles the
efficient allocation of buffers for intermediate results, it is important
to pay attention to such buffers in the design of the code. The first objective
is to avoid memory allocation in the middle of the training loop, i.e.,
all memory buffers should be allocated once and for all. Careless reuse
of the same memory buffers for different uses can however lead to bugs,
which can be checked, in the debugging phase,
by initializing buffers to the NaN (Not-A-Number) value,
which propagates into downstream computation (making it easy to
detect that uninitialized values were used)\footnote{~Personal communication
from David Warde-Farley, who learned this trick from Sam Roweis.}.

\subsection{Visualizations and Statistics}


The most basic statistics that should be measured during training are error statistics.
The {\em average loss} on the training set and the validation set and their evolution during
training are very useful to monitor progress and differentiate overfitting from
poor optimization. To make comparisons easier, it may be useful to compare neural
networks during training in terms of their ``age'' (number of updates made times 
mini-batch size $B$, i.e., number of examples visited) rather than in terms
of number of epochs (which is very sensitive to the training set size).

When using unsupervised training to learn the first few layers of a deep architecture,
a very common debugging and analysis tool is the {\em visualization of filters}, i.e.,
of the weight vectors associated with individual hidden units. This is simplest
in the case of the first layer and where the inputs are images (or image patches), time-series, or spectrograms
(all of which are visually interpretable). Several recipes have been proposed
to extend this idea to visualize the preferred input of hidden units in layers
that follow the first one~\citep{HonglakL2008-small,Erhan-vis-techreport-2010}.
In the case of the first layer, since one often obtains Gabor filters, a parametric
fit of these filters to the weight vector can be done so as to visualize the
distribution of orientations, positions and scales of the learned filters.
An interesting special case of visualizing first-layer weights is the
visualization of {\em word embeddings} (see Section~\ref{sec:symbols} below)
using a dimensionality reduction technique such as t-SNE~\citep{VanDerMaaten08-small}.

An extension of the idea of visualizing filters (which can apply to non-linear
or deeper features) is that of visualizing local (arount the given test point)
leading tangent vectors, i.e., the main directions in input space to which the
representation (at a given layer) is most sensitive to~\citep{Dauphin-et-al-NIPS2011-small}.

In the case where the inputs are not images or easily visualizable, or to get
a sense of the weight values in different hidden units, Hinton diagrams~\citep{Hinton89b}
are also very useful, using small squares whose color (black or white)
indicates a weight's sign and whose area represents its magnitude.

Another way to visualize what has been learned by an unsupervised (or joint
label-input) model is to look at samples from the model. Sampling
procedures have been defined at the outset for RBMs, Deep Belief Nets, and
Deep Boltzmann Machines, for example based on Gibbs sampling. When weights
become larger, mixing between modes can become very slow with Gibbs sampling.
An interesting
alternative is
rates-FPCD~\citep{TielemanT2009-small,Breuleux+Bengio-2011} which appears to 
be more robust to this problem and generally mixes faster, but at the cost of losing theoretical
guarantees. 

In the case of auto-encoder variants, it was not clear until
recently whether they were really capturing the underlying density (since they
are not optimized with respect to the maximum likelihood principle or an approximation
of it). It was therefore even less clear if there existed
appropriate sampling algorithms for auto-encoders, but a recent proposal 
for sampling from contractive auto-encoders appears to be working very well~\citep{Rifai-icml2012},
based on arguments about the geometric interpretation of the first derivative of the 
encoder~\citep{Bengio-arxiv-moments-2012}, showing that denoising and contractive
auto-encoders capture local moments (first and second) of the training density.

To get a sense of what individual hidden units represent, it has also
been proposed to vary only one unit while keeping the others fixed, e.g., to the
value obtained by finding the hidden units representation associated
with a particular input example.

Another interesting technique is the {\em visualization of the learning trajectory 
in function space}~\citep{Erhan+al-2010-small}. The idea is to associate the function
(as opposed to simply the parameters) computed by a neural network with
a low-dimensional (2-D or 3-D) representation, e.g., with the t-SNE ~\citep{VanDerMaaten08-small}
or Isomap~\citep{Tenenbaum2000-isomap} algorithms,
and then plot the evolution of
this function during training, or the population of such trajectories for
different initializations. This provides visualization of {\em effective local 
minima}\footnote{~It is
difficult to know for sure if it is a true local minima or if it appears like
one because the optimization algorithm is stuck.}
and shows that no two different random initializations
ended up in the same effective local minimum.

Finally, another useful type of visualization is to display statistics
(e.g., histogram, mean and standard deviation) of activations (inputs and
outputs of the non-linearities at each layer), activation gradients,
parameters and parameter gradients, by groups (e.g. different layers,
biases vs weights) and across training
iterations. See~\citet{GlorotAISTATS2010-small} for a practical example.  A
particularly interesting quantity to monitor is the discriminative ability
of the representations learnt at each layer, as discussed
in~\citep{Montavon-et-al-AISTATS2012}, and ultimately leading to an
analysis of the disentangled factors captured by the different layers as we
consider deeper architectures.


\section{Other Recommendations}
\label{sec:other}

\subsection{Multi-core machines, BLAS and GPUs}

Matrix operations are the most time-consuming in efficient implementations
of many machine learning algorithms and this is particularly true of neural
networks and deep architectures.  The basic operations are matrix-vector
products (forward propagation and back-propagation) and vector times vector
outer products (resulting in a matrix of weight gradients).
Matrix-matrix multiplications can be done substantially
faster than the equivalent sequence of matrix-vector products for two
reasons: by smart caching mechanisms such as implemented in the BLAS
library (which is called from many higher-level environments such as
python's numpy and Theano, Matlab, Torch or Lush), and thanks to
parallelism. Appropriate versions of BLAS can take advantage of multi-core
machines to distribute these computations on multi-core machines. The
speed-up is however generally a fraction of the total speedup one can hope
for (e.g. 4$\times$ on a 4-core machine), because of communication overheads and
because not all computation is parallelized. Parallelism becomes more
efficient when the sizes of these matrices is increased, which is why
mini-batch updates can be computationally advantageous, and more so when
more cores are present.

The extreme multi-core machines are the GPUs (Graphics Processing Units),
with hundreds of cores. Unfortunately, they also come with constraints and
specialized compilers which make it more difficult to fully take advantage
of their potential. On 512-core machines, we are routinely able to get
speed-ups of 4$\times$ to 40$\times$ for large neural networks. To make the use of GPUs
practical, it really helps to use existing libraries that efficiently
implement computations on GPUs. See~\citet{bergstra+al:2010-scipy-small} for
a comparative study of the Theano library (which compiles numpy-like
code for GPUs). One practical issue is that only the GPU-compiled
operations will typically be done on the GPU, and that transfers
between the GPU and CPU considerably slow things down. It is important
to use a profiler to find out what is done on the GPU and how
efficient these operations are in order to quickly invest one's
time where needed to make an implementation GPU-efficient and
keep most operations on the GPU card.

\subsection{Sparse High-Dimensional Inputs}

Sparse high-dimensional inputs can be efficiently handled by traditional
supervised neural networks by using a sparse matrix
multiplication. Typically, the input is a sparse vector while the weights
are in a dense matrix, and one should use an efficient implementation made
for just this case in order to optimally take advantage of sparsity. There
is still going to be an overhead on the order of 2$\times$ or more (on the
multiply-add operations, not the others) compared to a dense implementation
of the matrix-vector product.

For many unsupervised learning algorithms there is unfortunately a
difficulty. The computation for these learning algorithms usually involves
some kind of reconstruction of the input (like for all auto-encoder variants, but
also for RBMs and sparse coding variants), as if the inputs were in the
output space of the learner. Two exceptions to this problem are
semi-supervised embedding~\citep{WestonJ2008-small} and Slow Feature
Analysis~\citep{WisSej2002,BerkWisk2002a-small}.  The former pulls the
representation of nearby examples near each other and pushes dissimilar
points apart, while also tuning the
representation for a supervised learning task. The latter maximizes the
learned features' variance while minimizing their covariance and maximizing
their temporal auto-correlation. 

For algorithms that do need a form of
input reconstruction, an efficient approach based on {\em sampled
  reconstruction}~\citep{Dauphin+al-2011-small} has been proposed, successfully
implemented and evaluated for the case of auto-encoders and denoising
auto-encoders.  The first idea is that on each example (or mini-batch), one
samples a subset of the elements of the reconstruction vector, along with
the associated reconstruction loss. One only needs to compute the reconstruction
and the loss associated with these sampled
elements (or features), as well as the associated back-propagation operations into hidden
units and reconstruction weights. That alone would multiplicatively reduce the
computational cost by the amount of sparsity but make the gradient much more noisy and possibly
biased as well, if the sampling distribution was chosen not uniform. To reduce the
variance of that estimator, the idea is to guess for which features the
reconstruction loss will be larger and to sample with higher probability
these features (and their loss). In particular, the authors always sample
the features with a non-zero in the input (or the corrupted
input, in the denoising case), and uniformly sample an equal number of
those with a zero in the input and corrupted input. To make the estimator unbiased now requires
introducing a weight on the reconstruction loss associated with each
sampled feature, inversely proportional to the probability of sampling it,
i.e., this is an importance sampling scheme. The experiments show that the
speed-up increases linearly with the amount of sparsity while the average
loss is optimized as well as in the deterministic full-computation case.

\subsection{Symbolic Variables, Embeddings, Multi-Task Learning and Multi-Relational Learning}
\label{sec:symbols}

Parameter
sharing~\citep{Lang+Hinton88,LeCun89a,Lang+Hinton88,caruana93a-small,baxter95a-small,baxter97} is an
old neural network technique for increasing statistical power: if a
parameter is used in $N$ times more contexts (different tasks, different
parts of the input, etc.) then it may be as if we had $N$ times more
training examples for tuning its value. More examples to estimate a parameter reduces its variance (with
respect to sampling of training examples), which
is directly influencing generalization error: for example the generalization mean squared error can be
decomposed as the sum of a bias term and a variance term~\citep{Geman92}. The {\bf reuse}
idea was first exploited by applying the same parameter to
different parts of the input, as in convolutional neural
networks~\citep{Lang+Hinton88,LeCun89a}.
Reuse was also exploited by sharing the
lower layers of a network (and the representation of the input that they
capture) across multiple tasks associated with different outputs of the
network~\citep{caruana93a-small,baxter95a-small,baxter97}. This idea is also one of the
key motivations behind Deep Learning~\citep{Bengio-2009-book} because one
can think of the intermediate features computed in higher (deeper) layers
as different tasks that can share the sub-features computed in lower
layers (nearer the input). This very basic notion of reuse is
key to improving generalization in many settings, guiding the design
of neural network architectures in practical applications as well.

An interesting special case of these ideas is in the context of learning
with symbolic data. If some input variables are symbolic, taking value
in a finite alphabet, they can be represented as neural network inputs
by a one-hot subvector of the input vector (with a 0 everywhere except at the position
associated with the particular symbol). Now, sometimes
different input variables refer to different instances of the same {\em type}
of symbol. A patent example is with neural language 
models~\citep{Bengio-nnlm2003-small,Bengio-scholarpedia-2007}, where the input is
a sequence of words. In these models, the same input layer weights are reused
for words at different positions in the input sequence
(as in convolutional networks). The product of a one-hot sub-vector with
this shared weight matrix is a generally dense vector, and this associates
each symbol in the alphabet with a point in a vector space\footnote{~the result of the
matrix multiplication, which equals one of the columns of the matrix}, which
we call its {\em embedding}.
The idea of vector space representations for words and symbols is older~\citep{Deerwester90-small}
and is a particular case of the notion of {\em distributed representation}~\citep{Hinton86b-small,Hinton89b}
central to the connectionist approaches.
Learned embeddings of symbols (or other objects) can be conveniently visualized using
a dimensionality reduction algorithm such as t-SNE~\citep{VanDerMaaten08-small}.

In addition to sharing the embedding parameters across positions of words
in an input sentence, \citet{collobert:2011b} share them across natural
language processing tasks such as Part-Of-Speech tagging, chunking and
semantic role labeling.  Parameter sharing is a key idea behind
convolutional nets, recurrent neural networks and dynamic Bayes nets, in
which the same parameters are used for different temporal or spatial slices
of the data. This idea has been generalized from sequences and 2-D images
to arbitrary graphs with recursive neural networks or recursive graphical
models~\citep{Pollack90,Frasconi-1998,tr-bottou-2011,Socher-2011-small}, Markov
Logic Networks~\citep{Richardson+Domingos-2006} and
relational learning~\citep{Getoor+Taskar-book-2007}. A relational
database can be seen as a set of objects (or typed values) and relations between them,
of the form (object1, relation-type, object2). The same global set of parameters
can be shared to characterize such relations, across relations (which can be
seen as tasks) and objects. Object-specific parameters are the parameters
specifying the embedding of a particular discrete object.
One can think of the elements of each embedding vector as {\em implicit learned
  attributes}. Different tasks may demand different
attributes, so that objects which share some underlying characteristics and behavior
should end up having similar values of some of their attributes. For example,
words appearing in semantically and syntactically similar contexts end up getting
a very close embedding~\citep{collobert:2011b}.
If the same attributes can be useful for several tasks, then statistical power is gained
through parameter sharing, and transfer of information between tasks can happen, making
the data of some task informative for generalizing properly on another task.

The idea proposed
in \citet{bordes-aaai-2011,Antoine-al-2012-small} is to learn an energy function
that is lower for positive (valid) relations present in the training set, and
parametrized in two parts: on the one hand the symbol embeddings and on the
other hand the rest of the neural network that maps them to a scalar energy.
In addition, by considering relation types themselves as particular symbolic objects,
the model can reason about relations themselves and have relations 
between relation types. For example, `{\tt To be}' can act as a relation type
(in subject-attribute relations) but in the statement `` `{\tt To be}' is a verb''
it appears both as a relation type and as an object of the relation.

Such multi-relational learning opens the door to the application of neural networks
outside of their traditional applications, which was based on a single homogeneous source
of data, often seen as a matrix with one row per example and one column (or group of columns) per 
random variable. Instead, one often
has multiple heterogeneous sources of data (typically providing examples seen
as a tuple of values), each involving different random variables.
So long as these different sources {\em share some variables}, then the above multi-relational
multi-task learning approaches can be applied. Each variable can be associated
with its embedding function (that maps the value of a variable to a generic
representation space that is valid {\em across tasks and data sources}). This framework can
be applied not only to symbolic data but to mixed symbolic/numeric data if
the mapping from object to embedding is generalized from a table look-up to
a parametrized function (the simplest being a linear mapping) from its raw
attributes (e.g., image features) to its embedding. This has been exploited
successfully to design image search systems in which images and queries
are mapped to the same semantic space~\citep{Weston-ijcai2011}.

\section{Open Questions}
\label{sec:open}

\subsection{On the Added Difficulty of Training Deeper Architectures}

There are experimental results which provide some evidence that, at least
in some circumstances, deeper neural networks are more difficult to train
than shallow ones, in the sense that there is a greater chance of missing
out on better minima when starting from random initialization. This is borne
out by all the experiments where we find that some initialization scheme
can drastically improve performance. In the
Deep Learning literature this has been shown with the use of unsupervised
pre-training (supervised or not), both applied to supervised tasks --- training
a neural network for classification ~\citep{Hinton06-small,Bengio-nips-2006-small,ranzato-07-small} --- 
and unsupervised
tasks --- training a Deep Boltzmann Machine to model the data distribution~\citep{Salakhutdinov2009-small}.

The learning trajectories visualizations of~\citet{Erhan+al-2010-small} have shown
that even when starting from nearby configurations in function space, different
initializations seem to always fall in a different effective local minimum.
Furthermore, the same study showed that the minima found when using unsupervised
pre-training were far in function space from those found from random initialization,
in addition to giving better generalization error. Both of these findings highlight
the importance of initialization, hence of local minima effects, in deep
networks. Finally, it has been shown that these effects were both increased when
considering deeper architectures~\citep{Erhan+al-2010-small}.

There are also results showing that specific ways of setting the initial
distribution and ordering of examples (``curriculum learning'') can yield
better solutions~\citep{Elman93,Bengio+al-2009-small,Krueger+Dayan-2009}. This
also suggest that very particular ways of initializing parameters, very
different from uniformly sampled, can have a strong impact on the solutions
found by gradient descent. The hypothesis proposed in~\citep{Bengio+al-2009-small}
is that curriculum learning can act similarly to a {\em continuation method},
i.e., starting from an easier optimization task (e.g. convex) and tracking the local minimum
as the learning task is gradually made more difficult and closer to the
real task of interest.

Why would training deeper networks be more difficult? This is clearly still
an open question. A plausible partial answer is that deeper networks
are also more non-linear
(since each layer composes more non-linearity on top
of the previous ones), making gradient-based methods less efficient.
It may also be that the number and structure of local minima both change
qualitatively as we increase depth. Theoretical arguments 
support a potentially exponential gain in expressive power
of deeper architectures~\citep{Bengio-2009-book,Bengio+Delalleau-ALT-2011-small}
and it would be plausible that with this added expressive power
coming from the combinatorics of composed reuse of sub-functions
could come a corresponding increase in the number (and possibly
quality) of local minima. But the best ones could then also be more
difficult to find.

On the practical side, several experimental results point to factors
that may help training deep architectures:
\begin{itemize}
\item {\bf A local training signal.} What many successful procedures for
training deep networks have in common is that they involve a local training
signal that helps each layer decide what to do without requiring the
back-propagation of gradients through many non-linearities. This includes
of course the many variants of greedy layer-wise pre-training but
also the less well-known semi-supervised embedding algorithm~\citep{WestonJ2008-small}.
\item {\bf Initialization in the right range.} Based on the idea
that both activations and gradients should be able to flow well
through a deep architecture without significant reduction in variance,
\citet{GlorotAISTATS2010-small} proposed setting up the initial weights
to make the Jacobian of each layer have singular values near 1
(or preserve variance in both directions).
In their experiments this clearly helped greatly reducing the 
gap between purely supervised and pre-trained deep networks.
\item {\bf Choice of non-linearities}. In the same study~\citep{GlorotAISTATS2010-small}
and a follow-up~\citep{Glorot+al-AI-2011-small} it was shown that the choice of
hidden layer non-linearities interacted with depth. In particular, without
unsupervised pre-training, a deep neural network with sigmoids in the
top hidden layer would get stuck for a long time on a plateau and generally
produce inferior results, due to the special role of 0 and of the initial gradients
from the output units. Symmetric non-linearities like the hyperbolic
tangent did not suffer from that problem, while softer non-linearities 
(without exponential tails) such as the {\em softsign} function $s(a)=\frac{a}{1+|a|}$ worked
even better. In \citet{Glorot+al-AI-2011-small} it was shown that an asymmetric
but hard-limiting non-linearity such as the rectifier ($s(a)=\max(0,a)$, see also~\citep{Nair+Hinton-2010-small}) 
actually worked very well (but should not be used for output units), in spite
of the prior belief that the fact that when hidden units are saturated, gradients
would not flow well into lower layers. In fact gradients flow very well, but
on selected paths, possibly making the credit assignment (which parameters should
change to handle the current error) sharper and the Hessian condition number better. 
A recent heuristic that is related to the difficulty
of gradient propagation through neural net non-linearities is the 
idea of ``centering'' the non-linear operation such that each
hidden unit has zero average output and zero average slope~\citep{Schraudolph-1998,Raiko-2012-small}.
\end{itemize}

\subsection{Adaptive Learning Rates and Second-Order Methods}
\label{sec:adaptive}

To improve convergence and remove learning rates from the list of
hyper-parameters, many authors have advocated
exploring adaptive learning rate methods, either for a global learning rate~\citep{ICML2011Cho_98-small},
a layer-wise learning rate, a neuron-wise learning rate, or a parameter-wise
learning rate~\citep{bordes-09} (which then starts to look like a diagonal Newton method).
\citet{LeCun-these87,LeCun+98backprop-small} advocate the use of a second-order diagonal Newton
(always positive) approximation, with one learning rate per parameter (associated with
the approximated inverse second derivative of the loss with respect to
the parameter).
\citet{Hinton-RBMguide} proposes scaling learning rates so that the average
weight update is on the order of 1/1000th of the weight magnitude.
\citet{LeCun+98backprop-small} also propose a simple power method in order
to estimate the largest eigenvalue of the Hessian (which would be the
optimal learning rate). An interesting alternative to variants of
Newton's method are variants of the {\em natural gradient} method~\citep{amari98natural},
but like the basic Newton method it is computationally too expensive, requiring
operations on a too large square matrix (number of parameters by number of parameters).
Diagonal and low-rank online approximations of natural gradient~\citep{LeRoux+al-tonga-2008-small,LeRoux-chapter-2011-small}
have been proposed and shown to speed-up training in some contexts.
Several adaptive learning rate procedures have been proposed recently
and merit more attention and evaluations in the neural network
context, such as {\em adagrad}~\citep{Duchi+al-2011} and the adaptive learning
rate method from~\citet{Schaul2012} which claims to remove completely the
need for a learning rate hyper-parameter.

Whereas stochastic gradient descent converges very quickly initially it is
generally slower than second-order methods for the final convergence, and this may
be important in some applications.  As a consequence, batch training algorithms (performing only one update after
seeing the whole training set) such as the Conjugate Gradient method (a second
order method) have
dominated stochastic gradient descent for not too large datasets (e.g. less than thousands or
tens of thousands of examples).  Furthermore, it has recently been proposed and successfully
applied to use second-order methods over {\em large mini-batches}~\citep{Le-ICML2011-small,martens2010hessian-small}.
The idea is to do just a few iterations of the second-order methods on each mini-batch
and then move on to the next mini-batch, starting from the best previous point found.
A useful twist is to start training with one or more epoch of SGD, since SGD remains the
fastest optimizer early on in training.

At this point in time however, although the second-order and natural gradient 
methods are appealing conceptually,
have demonstrably helped in the studied cases
and may in the end prove to be very important, they have not yet
become a standard for neural networks optimization 
and need to be validated and maybe improved by other
researchers, before displacing simple (mini-batch) stochastic gradient
descent variants.


\subsection{Conclusion}

In spite of decades of experimental and theoretical work on artificial neural
networks, and with all the impressive progress made since the first edition of
this book, in particular in the area of Deep Learning,
there is still much to be done to better train neural networks and better
understand the underlying issues that can make the training task difficult. As stated
in the introduction, the wisdom distilled here should be taken as a guideline,
to be tried and challenged, not as a practice set in stone. The practice
summarized here, coupled with the increase in available computing power, 
now allows researchers to train neural networks on a scale that is far
beyond what was possible at the time of the first edition of this book,
helping to move us closer to artificial intelligence.

\subsubsection*{Acknowledgements}

The author is grateful for the comments and feedback provided by Nicolas
Le Roux, Ian Goodfellow, James Bergstra, Guillaume Desjardins, Razvan Pascanu, David
Warde-Farley, Eric Larsen, Frederic Bastien, and Sina Honari,
as well as for the financial support of NSERC, FQRNT, CIFAR, and the
Canada Research Chairs.

\bibliographystyle{natbib}
\bibliography{strings,ml,aigaion}

\end{document}